\documentclass[11pt]{article}

\usepackage[final]{acl}

\usepackage{times}
\usepackage{latexsym}

\usepackage[T1]{fontenc}

\usepackage[utf8]{inputenc}

\usepackage{microtype}

\usepackage{inconsolata}

\usepackage{graphicx}
\usepackage[most]{tcolorbox}

\usepackage{enumerate}
\usepackage{tabularx}
\usepackage[inline, shortlabels]{enumitem}

\usepackage{booktabs}
\usepackage[table]{xcolor}

\usepackage{wrapfig}   
\usepackage{caption}   

\usepackage{amssymb}

\newcommand{\dataset}{DUET} 

\title{Anatomy of Unlearning: The Dual Impact of Fact Salience and\\ Model Fine-Tuning}

\author{
  \textbf{Anna Borisiuk\textsuperscript{1, 4}\thanks{\texttt{a8or1suk@gmail.com}}},
  \textbf{Andrey Savchenko\textsuperscript{2,3,5}},
  \textbf{Alexander Panchenko\textsuperscript{1, 4}}, 
  \textbf{Elena Tutubalina\textsuperscript{1,5}}\\
  \textsuperscript{1}AIRI, 
  \textsuperscript{2}Sber AI Lab, 
  \textsuperscript{3}HSE University, 
  \textsuperscript{4}Skoltech, \\
  \textsuperscript{5}ISP RAS Research Center for Trusted Artificial Intelligence 
  }

\begin{document}
\maketitle

\begin{abstract}
Machine Unlearning (MU) enables Large Language Models (LLMs) to remove unsafe or outdated information. However, existing work assumes that all facts are equally forgettable and largely ignores whether the forgotten knowledge originates from pretraining or supervised fine-tuning (SFT). In this paper, we introduce the benchmark \dataset{} (Dual Unlearning Evaluation across Training Stages) composed of Wikidata-derived triplets annotated with fact popularity scores derived from Wikipedia link counts and LLM-based salience scores. Our experiments show that pretrained and SFT models respond differently to unlearning. An SFT step on the forget data yields smoother forgetting, more stable tuning, and 10-50\% higher retention, while direct unlearning for pretrained models remains unstable and prone to relearning or catastrophic forgetting. 
\end{abstract}

\section{Introduction}

\begin{figure}[t]
  \centering
  \makebox[\columnwidth][c]{%
    \includegraphics[width=1.\columnwidth]{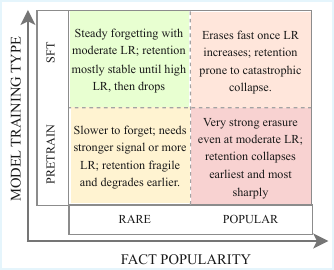}%
  }
  \caption{Unlearning landscape across fact popularity and model training type. Existing unlearning work does not account for popularity, implicitly assuming all facts are equal. Most studies evaluate forgetting on pretrained or SFT models without contrasting the two. The underexplored quadrant concerns analyzing how unlearning differs between pretrained and SFT models when fact popularity is taken into account.}
  \label{fig:mu_modes}
  \vspace{-0.5cm}
\end{figure}

\begin{figure*}[t]
    \centering
    \begin{minipage}[b]{0.33\textwidth}
        \centering
        \raisebox{20pt}{\includegraphics[width=\textwidth]{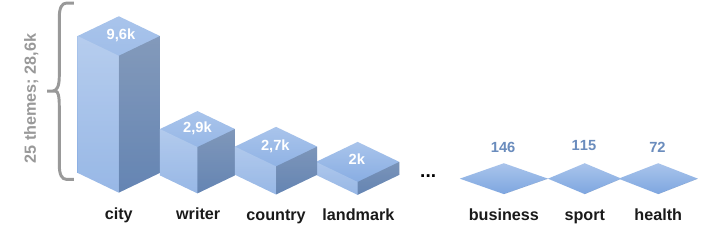}}
        \vspace{3pt}
        (a) distribution of topics
    \end{minipage}
    \hfill
    \begin{minipage}[b]{0.33\textwidth}
        \centering
        \includegraphics[width=\textwidth]{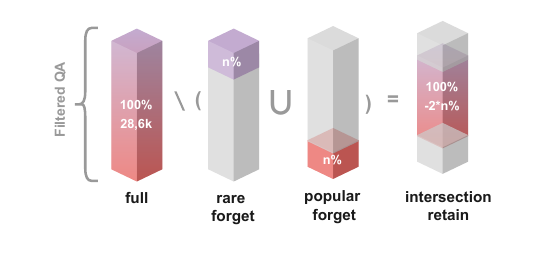}
        \vspace{3pt}
        (b) distribution of entity frequency
    \end{minipage}
    \hfill
    \begin{minipage}[b]{0.32\textwidth}
        \centering
        \includegraphics[width=\textwidth]{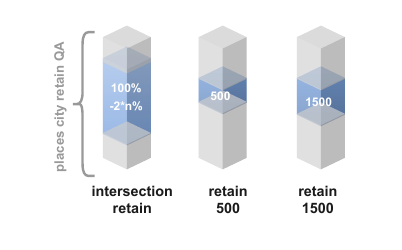}
        \vspace{3pt}
        (c) retain sets
    \end{minipage}
    \caption{
        \textbf{Overview of the \dataset{} benchmark.}
        (a) Topic distribution across 25 semantic themes (28.6k validated QA pairs), dominated by the \textit{places city} domain.
        (b) The filtering and stratification to derive rare and popular forget sets and the retain intersection.
        (c) Compact retain subsets (\textit{retain 500} and \textit{retain 1500}) drawn from \textit{places city}, providing resource-efficient yet structurally consistent evaluation settings.
    }
    \label{fig:\dataset{}_dataset}
    \vspace{-0.5cm}
\end{figure*}

Large Language Models have become central to modern NLP applications, yet their strong memorization of training data raises pressing questions about how to remove unsafe, outdated, or private information after deployment. Machine Unlearning aims to erase specific knowledge while preserving the model's overall competence, enabling safer and more adaptable systems~\cite{cao2015towards, sekhari2021remember, kurmanji2023towards, yuan2025towards}. Despite rapid progress, two fundamental aspects remain underexplored. First, prior work often assumes that all facts are equally forgettable, overlooking how knowledge frequency and real-world prominence affect persistence in model parameters. Popular facts, frequent and widely distributed, may be more deeply embedded than rare ones, making them harder to erase~\cite{wang2025selective, yuan2025towards}. Second, the role of the training paradigm is rarely systematically examined: most studies focus on supervisely fine-tuned (SFT) models built on synthetic data~\cite{tofu2024, shimuse}, while a few recent efforts~\cite{xu2025unlearning, li2024wmdp} explore pretrained checkpoints without a controlled comparison to SFT.  This gap motivates our research question:

\textit{How do fact popularity and model training type jointly influence machine unlearning performance?}

To answer this, we introduce \textbf{\dataset{}}, a benchmark of 28.6k Wikidata-derived question-answer pairs annotated for fact popularity using Wikipedia statistics and model-perceived salience. \dataset{} enables controlled comparison of unlearning performance across popularity levels and between \textit{Pretrain} and \textit{SFT} versions of the same architecture. We systematically analyze how these two factors interact and find that popular facts are substantially harder to erase, while pretrained and SFT models respond qualitatively differently to the same unlearning signals (cf. Figure~\ref{fig:mu_modes}). These findings highlight that unlearning is shaped not only by the chosen algorithm but also by where and how knowledge resides in the model.

The code\footnote{\url{https://github.com/Anya-wUw/DUET}} and the dataset\footnote{\url{https://huggingface.co/datasets/SwetieePawsss/DUET}} are available online.

\section{Related Work}
\textbf{Machine unlearning}  aims to selectively remove the impact of specific data points from a trained model without requiring complete retraining \cite{mantelero2013eu, cao2015towards, sekhari2021remember}. The objective is to obtain a model that behaves as if the forgotten data were never part of its training set. Early work explored MU in classical machine learning, while more recent research targets textual unlearning in LLMs \cite{sekhari2021remember, kurmanji2023towards}. A large body of studies focus on the LLaMA family-model \cite{yuan2025towards, maini2024tofu, wang2025selective, si2023knowledge}, which have become a standard testbed due to their effectiveness in controlled unlearning experiments \cite{maini2024tofu, yuan2025towards}. For this reason, our experiments also center on \href{https://huggingface.co/meta-llama/Llama-3.1-8B}{LLaMA-3.1-8B} variants.

\textbf{Unlearning benchmarks} were proposed to evaluate MU methods. Many of them rely on synthetic data, creating controlled but artificial scenarios. For example, TOFU \cite{maini2024tofu} introduces 200 fictitious author profiles and constructs forget sets at 1\%, 5\%, and 10\% scales, with the remainder used to measure retention. While valuable for privacy-like setups, its synthetic nature does not capture the complexity of real-world knowledge distributions. Other synthetic benchmarks include \cite{geng2025comprehensive, hu2024unlearning, si2023knowledge}.

Benchmarks based on real-world data address different objectives but largely ignore knowledge prevalence. For example, WMDP \cite{li2024wmdp} provides 3,668 multiple-choice questions to assess forgetting of hazardous knowledge, while MUSE \cite{shimuse} emphasizes a multifaceted evaluation that covers dimensions such as efficacy and efficiency. These resources are essential, yet their focus is orthogonal to the role of fact popularity.

\textbf{Limitations of existing datasets} stems from the fact that,
they typically assumes all facts  equally forgettable, overlooking how strongly frequent, widely distributed knowledge may be embedded compared to rare facts. Second, little attention has been given to the role of the training paradigm: many benchmarks rely on SFT models built on synthetic data, e.g., \cite{tofu2024}, as pretrained models alone are often not directly evaluable, while other works, e.g., \cite{xu2025unlearning}, attempt unlearning directly on pretrained models. Consequently, the joint effect of fact popularity and model training type has not been systematically studied.


\section{\dataset{} Benchmark}



\paragraph{Data Source}
We construct \textbf{\dataset{}} from 57k Wikidata-derived factual triplets spanning 25 semantic topics, including places, cities, human writers, countries, landmarks, industries, and health symptoms (Figure~\ref{fig:\dataset{}_dataset}a).
For each fact, we compute a \textit{popularity score} as the sum of Wikipedia sitelinks of its subject and object entities, capturing external prominence.

\paragraph{Filtering and QA Generation}
We remove instances where the same subject–relation pair yields multiple candidate answers to guarantee a unique answer per question, keeping only the answer associated with the most popular object.
Each remaining triplet is converted into a natural-language question–answer pair (\textit{subject–relation → object}), following a similar disambiguation procedure as in~\cite{huang2025halluedit}.
To verify that the knowledge is already accessible to pretrained models, we prompt \textit{LLaMA-3.1-8B} and retain only 28.6k triplets where the model's generated answer reaches BERT cosine similarity above 0.6 with the gold answer, ensuring that unlearning targets facts the model has actually internalized rather than knowledge it never encoded in the first place.

\paragraph{Popularity Validation}
To confirm that Wikipedia-based popularity correlates with model-internal salience, we compare it with factual judgments from \textit{LLaMA-3.3-70B}.
The model rates each fact's prominence on a three-point scale: -1 (disagree), 0 (unknown), and 1 (agree).
The two signals are strongly correlated (80.65\% on 5\% forget splits and 81.95\% on the \textit{places city} subset; see Figure~\ref{fig:OR_popularity}), validating the reliability of our external metric.
Among existing QA datasets, the closest in spirit is \textbf{PopQA}~\citep{mallen2023not}, which also quantifies factual popularity; however, DUET is roughly twice as large and specifically designed for controlled unlearning experiments across training stages.

\paragraph{Popularity-Based Splits}
Following methodology from~\cite{tofu2024, dontsov-etal-2025-clear}, we stratify \dataset{} into unlearning tasks at 1\%, 5\%, and 10\% scales.
For a given proportion $N$, we define three complementary subsets (Figure~\ref{fig:\dataset{}_dataset}b):
\begin{enumerate}[topsep=1pt,itemsep=1pt,parsep=0pt,leftmargin=1.3em]
    \item \textbf{Rare forget set} is composed of bottom $N$\% of least popular facts.
    \item \textbf{Popular forget set} is composed of top $N$\% of most popular facts.
    \item \textbf{Retain intersection set} is composed of the remaining $(100 - 2N)$\% of data, used to evaluate the preservation of unaffected knowledge.
\end{enumerate}

\paragraph{City Sets}
To support efficient experimentation, we replicate the same popularity-based partitioning (rare vs. popular for forget and retain) within the most prominent domain, \textit{places city} (9.6k samples).
We construct smaller sets for rapid validation from this category: \textit{retain 500} and \textit{retain 1500}.
These subsets are randomly sampled and filtered with a stricter BERT similarity threshold ($>0.7$),
selected empirically to balance coverage and factual confidence: lower thresholds admit noisy
paraphrases, while higher thresholds excessively reduce set size. This ensures full alignment
with the structure of the complete benchmark (Figure~\ref{fig:\dataset{}_dataset}c).

\section{Experiments}
\label{sec:experiments}
\subsection{Experimental Setup}

\begin{figure*}[t]
    \centering
    \includegraphics[width=\textwidth]{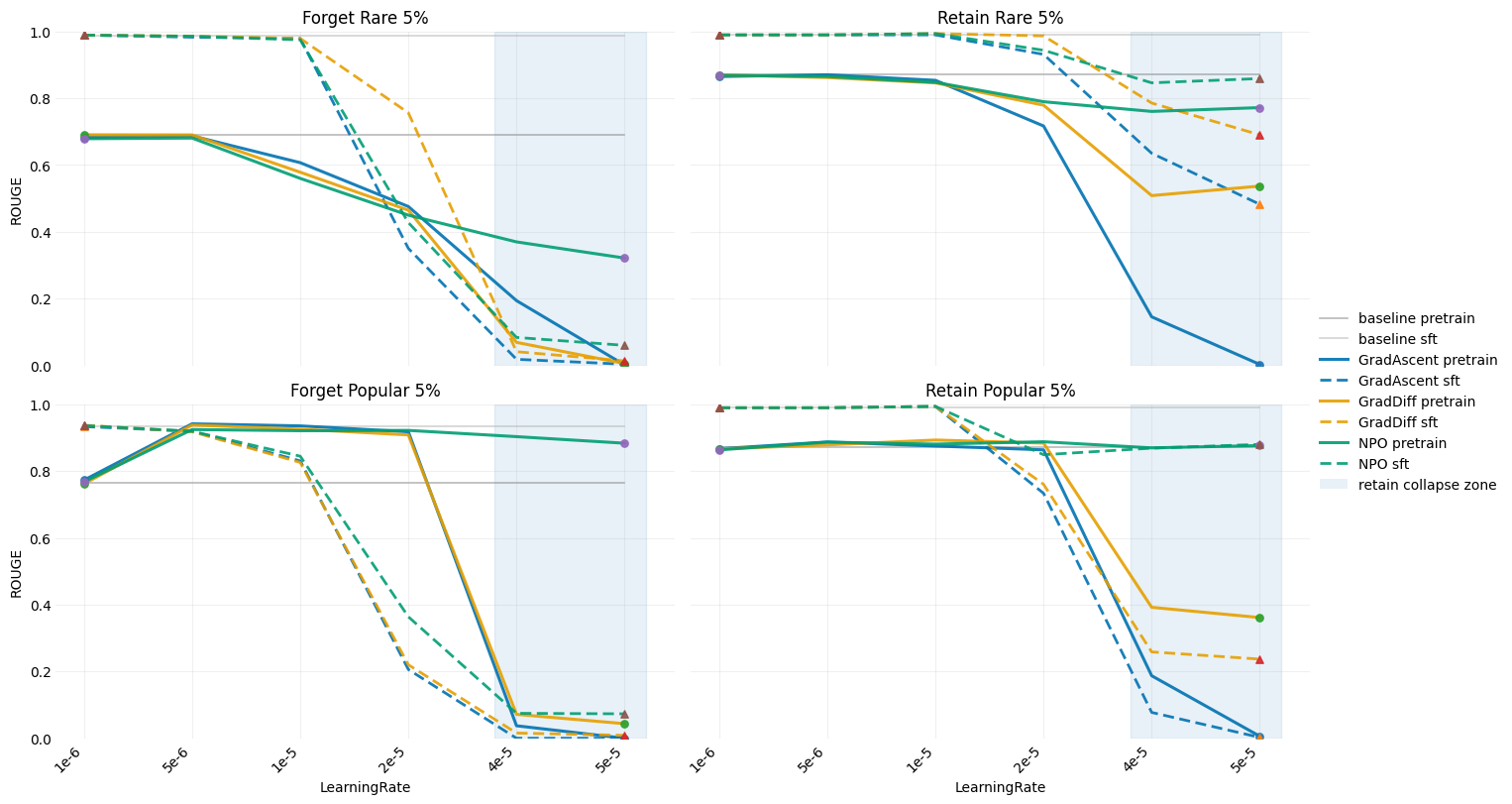}
    \caption{\textbf{Unlearning results for the city domain} and, top $N$ = 5\% rare/popular entities. Top: rare; bottom: popular. Left: forget (\texttt{city\_forget\_\{rare,popular\}\_5}); right: retain (\texttt{city\_fast\_retain\_500}). ROUGE is reported across learning rates. Pretrain is shown as a solid line with circles; SFT as a dashed line with stars. 
    }
    \label{fig:rare_forget_5}
\end{figure*}

We conduct our experiments using the LLaMA-3.1-8B model as the base architecture.
To obtain comparable Pretrained and SFT variants under identical conditions, we use the released checkpoint as the Pretrained model and train an SFT variant on the full \dataset{} dataset (28.6k samples) using LoRA. Parameter-efficient fine-tuning achieves performance close to full-model training while requiring substantially fewer resources~\citep{hu2021lora, compare2023full_lora, lin2023parameter_efficient}, and LoRA-based unlearning has emerged as the standard approach for large models~\citep{cha2025robust_unlearning, liu2025lune}.

We apply LoRA to the unlearning step as well. Full-parameter SFT produces substantially less stable unlearning: on the city-forget split with NPO at learning rate of $2\cdot 10^{-5}$, the full-SFT model retains a ROUGE-L of 0.997 on the forget set, while LoRA reduces it to 0.364. On multi-domain data, full-SFT models collapse catastrophically across all algorithms, dropping ROUGE-L to 0-0.14 on the forget set, whereas LoRA models maintain stable forgetting and near-unchanged retention. Full results are in Appendix~\ref{sec:appendix_full_sft}.

For unlearning experiments, we focus on the largest topical category, {places city} (9.6k samples), and apply three widely used algorithms: Gradient Ascent (GA)~\cite{jang2022knowledge}, Gradient Difference (GD)~\cite{liu2022continual}, and Negative Preference Optimization (NPO)~\cite{zhangnegative}.
We perform a grid search over learning rates from $10^{-6}$ to $5 \cdot 10^{-5}$ and training epochs from 1 to 3.
Learning rates above $5\cdot 10^{-5}$ consistently led to catastrophic forgetting, while rates below $10^{-6}$ resulted in negligible unlearning.
We identify two epochs as a practical compromise (cf. Figure~\ref{fig:rare_forget_5}): 1 epoch yields insufficient forgetting (cf. Figure~\ref{fig:rare_forget_1}), while three epochs cause excessive knowledge degradation.
We report primary results for $N$ of 5\%; experiments for $N$ of 1\% and 10\% are provided in Appendix~\ref{sec:appendix_forget_size} and show consistent qualitative trends.

We use the \textit{fast retain} subset of 500 samples to evaluate retention quality.
ROUGE-L is the primary metric for forgetting and retention; we additionally validate it via an LLM-as-a-Judge protocol (DeepSeek v1), scoring \textit{Accuracy} (factual alignment with the reference) and \textit{Fluency} (linguistic naturalness). Both metrics are reported for LLaMA in Table~\ref{tab:forget_retain_llama}; Gemma results are in Appendix~\ref{subsec:llm_judge}.

\begin{table*}[t]
\centering
\small
\setlength{\tabcolsep}{4pt}
\renewcommand{\arraystretch}{1.12}
\begin{tabular}{lll rr cc cc}
\toprule
& & & \multicolumn{2}{c}{\textbf{ROUGE-L}} & \multicolumn{2}{c}{\textbf{Judge Acc.}} & \multicolumn{2}{c}{\textbf{Judge Flu.}} \\
\cmidrule(lr){4-5}\cmidrule(lr){6-7}\cmidrule(lr){8-9}
\shortstack{\textbf{Train}\\\textbf{type}} & \textbf{Algo} & \textbf{Pop.} &
Forget $\downarrow$ & Retain $\uparrow$ & Forget $\downarrow$ & Retain $\uparrow$ & Forget $\uparrow$ & Retain $\uparrow$ \\
\midrule
Pretrain & w/o & rare & 0.691 & 0.871 & 0.584 & 0.839 & 0.701 & 0.840 \\
\rowcolor{gray!10} SFT & w/o & rare & 0.987 & 0.990 & 0.773 & 0.862 & 0.741 & 0.649 \\
Pretrain & GA  & rare & 0.476 {\scriptsize \textcolor{green!50!black}{($-$0.22)}} & 0.717 {\scriptsize \textcolor{red!65!black}{($-$0.15)}} & 0.368 {\scriptsize \textcolor{green!55!black}{($-$0.22)}} & 0.670 {\scriptsize \textcolor{red!65!black}{($-$0.17)}} & 0.646 {\scriptsize \textcolor{red!65!black}{($-$0.06)}} & 0.748 {\scriptsize \textcolor{red!65!black}{($-$0.09)}} \\
\rowcolor{gray!10} SFT & GA  & rare & 0.350 {\scriptsize \textcolor{green!50!black}{($-$0.64)}} & 0.931 {\scriptsize \textcolor{red!65!black}{($-$0.06)}} & 0.507 {\scriptsize \textcolor{green!55!black}{($-$0.27)}} & 0.842 {\scriptsize \textcolor{red!65!black}{($-$0.02)}} & 0.642 {\scriptsize \textcolor{red!65!black}{($-$0.10)}} & 0.615 {\scriptsize \textcolor{red!65!black}{($-$0.03)}} \\
Pretrain & GD  & rare & 0.465 {\scriptsize \textcolor{green!50!black}{($-$0.23)}} & 0.779 {\scriptsize \textcolor{red!65!black}{($-$0.09)}} & 0.358 {\scriptsize \textcolor{green!55!black}{($-$0.23)}} & 0.747 {\scriptsize \textcolor{red!65!black}{($-$0.09)}} & 0.577 {\scriptsize \textcolor{red!65!black}{($-$0.12)}} & 0.627 {\scriptsize \textcolor{red!65!black}{($-$0.21)}} \\
\rowcolor{gray!10} SFT & GD  & rare & 0.756 {\scriptsize \textcolor{green!50!black}{($-$0.23)}} & 0.987 {\scriptsize \textcolor{red!65!black}{($-$0.00)}} & 0.572 {\scriptsize \textcolor{green!55!black}{($-$0.20)}} & 0.853 {\scriptsize \textcolor{green!55!black}{($-$0.01)}} & 0.694 {\scriptsize \textcolor{red!65!black}{($-$0.05)}} & 0.612 {\scriptsize \textcolor{red!65!black}{($-$0.04)}} \\
Pretrain & NPO & rare & 0.450 {\scriptsize \textcolor{green!50!black}{($-$0.24)}} & 0.790 {\scriptsize \textcolor{red!65!black}{($-$0.08)}} & 0.349 {\scriptsize \textcolor{green!55!black}{($-$0.23)}} & 0.745 {\scriptsize \textcolor{red!65!black}{($-$0.09)}} & 0.608 {\scriptsize \textcolor{red!65!black}{($-$0.09)}} & 0.618 {\scriptsize \textcolor{red!65!black}{($-$0.22)}} \\
\rowcolor{gray!10} SFT & NPO & rare & 0.428 {\scriptsize \textcolor{green!50!black}{($-$0.56)}} & 0.944 {\scriptsize \textcolor{red!65!black}{($-$0.05)}} & 0.521 {\scriptsize \textcolor{green!55!black}{($-$0.25)}} & 0.828 {\scriptsize \textcolor{red!65!black}{($-$0.03)}} & 0.637 {\scriptsize \textcolor{red!65!black}{($-$0.10)}} & 0.589 {\scriptsize \textcolor{red!65!black}{($-$0.06)}} \\
\midrule
Pretrain & w/o & pop & 0.766 & 0.871 & 0.915 & 0.839 & 0.899 & 0.840 \\
\rowcolor{gray!10} SFT & w/o & pop & 0.935 & 0.990 & 0.862 & 0.862 & 0.606 & 0.649 \\
Pretrain & GA  & pop & 0.918 {\scriptsize \textcolor{red!65!black}{($+$0.15)}} & 0.864 {\scriptsize \textcolor{red!65!black}{($-$0.01)}} & 0.804 {\scriptsize \textcolor{green!55!black}{($-$0.11)}} & 0.752 {\scriptsize \textcolor{red!65!black}{($-$0.09)}} & 0.742 {\scriptsize \textcolor{red!65!black}{($-$0.16)}} & 0.720 {\scriptsize \textcolor{red!65!black}{($-$0.12)}} \\
\rowcolor{gray!10} SFT & GA  & pop & 0.206 {\scriptsize \textcolor{green!50!black}{($-$0.73)}} & 0.733 {\scriptsize \textcolor{red!65!black}{($-$0.26)}} & 0.261 {\scriptsize \textcolor{green!55!black}{($-$0.60)}} & 0.555 {\scriptsize \textcolor{red!65!black}{($-$0.31)}} & 0.604 {\scriptsize \textcolor{red!65!black}{($-$0.00)}} & 0.673 {\scriptsize \textcolor{green!55!black}{($+$0.02)}} \\
Pretrain & GD  & pop & 0.908 {\scriptsize \textcolor{red!65!black}{($+$0.14)}} & 0.884 {\scriptsize \textcolor{green!50!black}{($+$0.01)}} & 0.817 {\scriptsize \textcolor{green!55!black}{($-$0.10)}} & 0.760 {\scriptsize \textcolor{red!65!black}{($-$0.08)}} & 0.777 {\scriptsize \textcolor{red!65!black}{($-$0.12)}} & 0.753 {\scriptsize \textcolor{red!65!black}{($-$0.09)}} \\
\rowcolor{gray!10} SFT & GD  & pop & 0.220 {\scriptsize \textcolor{green!50!black}{($-$0.72)}} & 0.760 {\scriptsize \textcolor{red!65!black}{($-$0.23)}} & 0.254 {\scriptsize \textcolor{green!55!black}{($-$0.61)}} & 0.588 {\scriptsize \textcolor{red!65!black}{($-$0.27)}} & 0.644 {\scriptsize \textcolor{green!55!black}{($+$0.04)}} & 0.669 {\scriptsize \textcolor{green!55!black}{($+$0.02)}} \\
Pretrain & NPO & pop & 0.922 {\scriptsize \textcolor{red!65!black}{($+$0.16)}} & 0.888 {\scriptsize \textcolor{green!50!black}{($+$0.02)}} & 0.793 {\scriptsize \textcolor{green!55!black}{($-$0.12)}} & 0.823 {\scriptsize \textcolor{green!55!black}{($-$0.02)}} & 0.763 {\scriptsize \textcolor{red!65!black}{($-$0.14)}} & 0.758 {\scriptsize \textcolor{red!65!black}{($-$0.08)}} \\
\rowcolor{gray!10} SFT & NPO & pop & 0.364 {\scriptsize \textcolor{green!50!black}{($-$0.57)}} & 0.849 {\scriptsize \textcolor{red!65!black}{($-$0.14)}} & 0.276 {\scriptsize \textcolor{green!55!black}{($-$0.59)}} & 0.683 {\scriptsize \textcolor{red!65!black}{($-$0.18)}} & 0.703 {\scriptsize \textcolor{green!55!black}{($+$0.10)}} & 0.690 {\scriptsize \textcolor{green!55!black}{($+$0.04)}} \\
\bottomrule
\end{tabular}
\caption{LLaMA-3.1-8B at $N=5\%$, $lr=2\times10^{-5}$. ROUGE-L measures lexical forgetting and retention. Judge Accuracy measures factual alignment with the reference answer; Judge Fluency measures linguistic naturalness (DeepSeek v1). Parentheses show change from the matching \textit{w/o} baseline.}
\label{tab:forget_retain_llama}
\end{table*}

\subsection{Key Findings}

\paragraph{Finding 1: Pretrained and SFT models exhibit opposite behavior on popular facts} When unlearning {popular entities}, the Pretrained model displays counter-intuitive behavior: instead of forgetting, it {improves} performance on the forget set, with ROUGE-L scores increasing across all tested epochs (1-3). This suggests the model treats unlearning signals as additional fine-tuning on familiar knowledge. In stark contrast, the SFT model behaves as expected: unlearning consistently decreases ROUGE-L on the forget set. This divergence persists regardless of epoch count, indicating a fundamental difference in how the two model training types process unlearning interventions on well-known facts.

{For rare entities}, both models behave conventionally: forgetting metrics decline exponentially with training. As learning rates increase, both architectures eventually reach catastrophic forgetting thresholds where performance collapses uniformly.

\begin{tcolorbox}[colback=gray!10, colframe=black!50,
  boxrule=0.5pt, arc=3pt, left=6pt, right=6pt, top=4pt, bottom=4pt]
\textbf{Takeaway 1:} For \textit{popular entities}, a preliminary SFT step enables more stable and reliable unlearning.
\end{tcolorbox}

\paragraph{Finding 2: retention degradation differs dramatically by fact popularity} On the retain set, Pretrained models show relatively stable ROUGE-L scores regardless of whether rare or popular facts are removed: popularity has little effect on retained knowledge quality, though forgetting is accompanied by a sharp drop once catastrophic thresholds are reached.

In contrast, SFT models display the opposite trend: unlearning {popular} facts leads to catastrophic forgetting of the retain set, while unlearning {rare} facts does not cause such drastic quality loss. Thus, SFT models are overall more robust, reducing the risk of catastrophic retention collapse by roughly a factor of two compared to rare-fact forgetting. Pretrained models, by comparison, behave more uniformly across fact types but are prone to sudden quality drops when forgetting escalates.

\begin{tcolorbox}[colback=gray!10, colframe=black!50,
  boxrule=0.5pt, arc=3pt, left=6pt, right=6pt, top=4pt, bottom=4pt]
\textbf{Takeaway 2:} \textit{SFT models are more robust} on the retain set,
showing roughly half the risk of catastrophic forgetting compared to rare-fact removal,
while pretrained models degrade more abruptly regardless of fact type.
\end{tcolorbox}

Unlearning preserves overall capability: the worst-case deviation on MMLU and HellaSwag does not exceed 3\% across all tested configurations (cf. Appendix~\ref{subsec:llm_benchs} and Figure~\ref{fig:llm_benchs}). Representation-level analyses in Appendix~\ref{subsec:intrinsic} reveal that rare and popular facts respond asymmetrically to unlearning at the token and hidden-state levels: popular facts sustain higher token probability under gradient-based pressure, while rare facts show larger shifts in hidden-state similarity after unlearning.

Thus, the popularity of the fact and the type of model training jointly determine unlearning behavior. Pretrained models respond uniformly but abruptly to unlearning signals, with narrow and unstable hyperparameter windows that make reliable forgetting difficult to achieve. SFT models are strongly shaped by fact popularity: popular facts resist forgetting at low learning rates and cause larger retention drops at high ones, while rare facts are removed smoothly across the full range. Despite this sensitivity, SFT models exhibit substantially better forgetting dynamics overall, with retention quality 10 to 50\% higher than pretrained counterparts at the same learning rate (see Figure~\ref{fig:rare_forget_5}).

Both effects persist under distillation-based unlearning: UNDIAL experiments (cf. Appendix~\ref{sec:appendix_undial}) show that SFT retains near-baseline retention while forgetting progresses smoothly, and the rare/popular asymmetry holds under a self-distillation mechanism, confirming that these patterns are not tied to gradient-based objectives alone. These findings underscore the need to jointly evaluate data composition and model training type and highlight the practical benefit of a preliminary SFT step for achieving more stable, controllable unlearning.

We further evaluate the same setup on Gemma-7B and Qwen-2.5-7B to test model-specific effects.
As detailed in Appendix~\ref{add_experiment}, Gemma shows the same qualitative patterns as LLaMA: rare facts are easier to forget, while SFT variants yield smoother but more sensitive forgetting dynamics.
Multi-domain experiments in Appendix~\ref{subsec:multidomain} further confirm that the rare/popular asymmetry is not an artifact of the city-heavy distribution, but holds across balanced domain subsets.

\section{Conclusion}


We introduced  a Wikidata-based benchmark  annotated with fact popularity. It enabled us to conduct  experiments, which uncovered that unlearning dynamics are jointly depend on fact popularity and model training type.
Besides, we found that pretrained models are unstable, prone to abrupt degradation, and even unintended relearning of popular facts.
By contrast, SFT models provide a smoother forgetting, more reliable hyperparameter tuning, and up to 10--50\% higher retention quality.

These results challenge the common assumption that all facts are equally forgettable and that model training type is irrelevant. We argue that future MU methods must consider both the composition of the forget set and the origin of knowledge in the model. The proposed dataset offers enables a more reliable evaluation and a deeper understanding of how popularity and training paradigm interact in unlearning.

\section*{Limitations}
Our claims hold under the following boundary conditions. 
(i) Scope: experiments target LLaMA-3.1-8B and the \textit{Places City} forget set with a compact retain set; broader domains and larger models are future work. 
(ii) Popularity labels: we rely on Wikipedia signals and model salience; these proxies may drift over time and across languages. 
(iii) Metrics: we report ROUGE-L for free-form answers; complementary factuality and safety judgments, including human evaluation, are planned follow-ups. 
These limits do not alter the central result that popularity and training regime jointly shape unlearning outcomes.

\section*{Ethics Statement}
Our data come from public sources (Wikidata, Wikipedia). We do not collect sensitive attributes, and no human subjects were involved. We view machine unlearning as a contribution to AI safety and data governance because it enables the removal of unsafe, outdated, or private content from deployed models. We also used ChatGPT 5 for minor language and grammatical edits; all research design, analysis, and interpretation were conducted by the authors.

\section*{Acknowledgments}
The work was supported by a grant, provided by the Ministry of Economic Development of the Russian Federation in accordance with the subsidy agreement (agreement identifier 000000C313925P4G0002) and the agreement with the Ivannikov Institute for System Programming of the Russian Academy of Sciences dated June 20, 2025 No. 139-15-2025-011.


\appendix

\label{sec:appendix}

\section{Machine Unlearning: Definition and Algorithms}

\subsection{Problem Definition}

MU aims to remove the influence of specific data from a trained model without retraining it from scratch.
Given a language model $f_{\theta}$ with parameters $\theta$ and a dataset $\mathcal{D} = \{(q,a)\}$ of question–answer pairs, the goal is to make the model forget the knowledge contained in a subset of data, while keeping its general capabilities intact.
Formally, we define two disjoint subsets:
\begin{itemize}
    \item the \textbf{forget set} $\mathcal{D}_f \subset \mathcal{D}$ containing samples that should be unlearned;
    \item the \textbf{retain set} $\mathcal{D}_r = \mathcal{D} \setminus \mathcal{D}_f$ containing samples whose knowledge should be preserved.
\end{itemize}

The desired outcome is a new model $f_{\hat{\theta}}$ such that
\[
f_{\hat{\theta}}(q) \not\approx a, \quad \forall (q,a)\in\mathcal{D}_f,
\]
\[
f_{\hat{\theta}}(q) \approx a, \quad \forall (q,a)\in\mathcal{D}_r.
\]
In other words, the model should selectively erase information about the forget set while maintaining its performance on the retain set.

\subsection{Task Formulation}

The unlearning process can be expressed as a selective optimization problem that updates model parameters from $\theta$ to $\hat{\theta}$ by jointly enforcing two objectives:
(i) degrade performance on the forget set, and
(ii) preserve performance on the retain set.
This trade-off can be formalized as
\begin{equation}
    \min_{\theta}
    \Big[
      -\,\mathbb{E}_{(q,a)\in\mathcal{D}_f}
        L(a|q;\theta)
      +
      \lambda\,\mathbb{E}_{(q,a)\in\mathcal{D}_r}
        L(a|q;\theta)
    \Big],
\end{equation}
where $L$ is a loss function such as the negative log-likelihood, and $\lambda$ controls the balance between forgetting and retention.
This unified objective underlies most gradient-based MU methods and defines the optimization setting used in our experiments.

\subsection{Unlearning Algorithms}

\paragraph{Gradient Ascent (GA)}~\cite{jang2022knowledge}.
A basic approach that reverses standard training by maximizing the loss on the forget set $\mathcal{D}_f$.
It can be written as
\begin{equation}
\mathcal{L}^{\text{GA}}(\theta)
  = -\,\mathcal{L}_{\text{NLL}}(\mathcal{D}_f, \theta),
\end{equation}
where
\[
\mathcal{L}_{\text{NLL}}(\mathcal{D}, \theta)
  = \tfrac{1}{|\mathcal{D}|}
    \sum_{(q,a)\in\mathcal{D}}
    -\log P_{\theta}(a|q).
\]

\paragraph{Gradient Difference (GD)}~\cite{liu2022continual}.
An extension of GA that also preserves knowledge in the retain set.
It increases loss on $\mathcal{D}_f$ while decreasing it on $\mathcal{D}_r$:
\begin{equation}
\mathcal{L}^{\text{GD}}(\theta)
  = -\,\mathcal{L}_{\text{NLL}}(\mathcal{D}_f, \theta)
    + \lambda\,\mathcal{L}_{\text{NLL}}(\mathcal{D}_r, \theta).
\end{equation}

\paragraph{Negative Preference Optimization (NPO)}~\cite{zhangnegative}.
This method frames unlearning as a preference optimization problem, penalizing the model for assigning a higher likelihood to the forgotten answers relative to a reference model $\theta_{\text{ref}}$:
\begin{equation}
\resizebox{0.85\linewidth}{!}{$
\mathcal{L}^{\mathrm{NPO}}(\theta)
  = \tfrac{2}{\beta}\,
    \mathbb{E}_{(q,a)\in\mathcal{D}_f}
    \!\left[\log\!\left(1+\!\left(
    \tfrac{P_{\theta}(a\mid q)}{P_{\theta_{\mathrm{ref}}}(a\mid q)}\right)^{\!\beta}\right)\right],
$}
\end{equation}
where $\beta=1$ is used as recommended in the original formulation.

Each algorithm represents a different trade-off between erasing specific knowledge and preserving the rest, forming the basis for our comparative study in Section~\ref{sec:experiments}.

\section{Additional Experiments}

\begin{table*}[t]
\centering
\footnotesize
\caption{Examples of \dataset{} question--answer pairs with popularity annotations.}
\label{tab:qa_examples}
\begin{tabular}{|p{4.5cm}|p{3cm}|p{1.5cm}|p{1.2cm}|}
\hline
\textbf{Question} & \textbf{Answer} & \textbf{Pop. score} & \textbf{Label} \\
\hline
What is the founded by Welch's? & Thomas Bramwell Welch & 44 & rare \\
\hline
What is the instance of Viltolarsen? & antisense oligonucleotide & 47 & rare \\
\hline
What is the subclass of cowpunk? & punk music & 47 & rare \\
\hline
What is the capital of Poland? & Warsaw & 2435 & popular \\
\hline
What is the highest judicial authority of Australia? & High Court of Australia & 2443 & popular \\
\hline
What is the country of Cayenne? & France & 2445 & popular \\
\hline
\end{tabular}%
\end{table*}

\begin{figure*}[t]
    \centering
    \includegraphics[width=1.\textwidth]{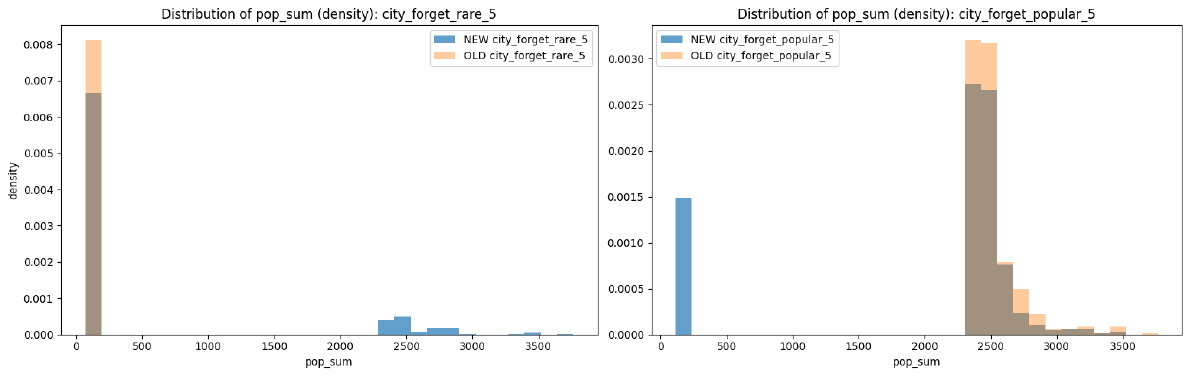}
    \caption{Distributions of popularity scores (\textit{pop\_sum}) for the \textit{city} domain in the new and old benchmark splits.
    The left panel shows the \textit{rare forget 5\%} subset and the right panel the \textit{popular forget 5\%} subset.
    Of 482 samples, 87 differ from the previous version, yielding an overall overlap of 81.95\%.}
    \label{fig:OR_popularity}
\end{figure*}

\subsection{Full-parameter SFT vs.\ LoRA SFT}
\label{sec:appendix_full_sft}

We compare LoRA-based SFT (used throughout the paper) against full-parameter SFT to rule out the possibility that LoRA artificially facilitates unlearning. In both settings, the unlearning step uses LoRA. Tables~\ref{tab:full_sft_city} and~\ref{tab:full_sft_multi} report ROUGE-L before and after unlearning at learning rates $2\times10^{-5}$, $4\times10^{-5}$, and $5\times10^{-6}$ for the city-forget split, and at $2\times10^{-5}$ for multi-domain data.

On the city-forget split (Table~\ref{tab:full_sft_city}), full-parameter SFT shows qualitatively similar trends to LoRA SFT: higher learning rates produce stronger forgetting. However, forgetting is consistently less effective. For example, NPO at $lr=2\times10^{-5}$ achieves a forget ROUGE-L of 0.997 on popular facts under full-parameter SFT, compared to 0.364 under LoRA SFT. On multi-domain data (Table~\ref{tab:full_sft_multi}), all three algorithms applied to the full-parameter SFT model result in near-complete output degradation, whereas LoRA SFT retains near-baseline retention throughout (see Table~\ref{tab:forget_retain_llama} and Figure~\ref{fig:rare_forget_5}). \textit{Conclusion.} LoRA SFT followed by LoRA unlearning is strictly more stable than full-parameter SFT: it achieves stronger forget-set reduction at the same learning rates and avoids the catastrophic multi-domain collapse observed in the full-parameter setting.

\begin{table}[t]
\centering
\footnotesize
\caption{Full-parameter SFT with LoRA unlearning on the city-forget split. ROUGE-L is reported before and after unlearning.}
\label{tab:full_sft_city}
\setlength{\tabcolsep}{4pt}
\resizebox{\columnwidth}{!}{%
\begin{tabular}{llllcc}
\toprule
Algorithm & LR & Popularity & ROUGE-L Before & ROUGE-L After \\
\midrule
GradAscent & 2e-5 & popular & 0.998 & 0.439 \\
GradAscent & 4e-5 & popular & 0.998 & 0.008 \\
GradAscent & 5e-6 & popular & 0.998 & 0.998 \\
GradAscent & 2e-5 & rare    & 0.994 & 0.991 \\
GradAscent & 4e-5 & rare    & 0.994 & 0.369 \\
GradAscent & 5e-6 & rare    & 0.994 & 0.993 \\
\midrule
GradDiff   & 2e-5 & popular & 0.998 & 0.821 \\
GradDiff   & 4e-5 & popular & 0.998 & 0.044 \\
GradDiff   & 5e-6 & popular & 0.998 & 0.998 \\
GradDiff   & 2e-5 & rare    & 0.994 & 0.993 \\
GradDiff   & 4e-5 & rare    & 0.994 & 0.942 \\
GradDiff   & 5e-6 & rare    & 0.994 & 0.993 \\
\midrule
NPO        & 2e-5 & popular & 0.998 & 0.997 \\
NPO        & 4e-5 & popular & 0.998 & 0.214 \\
NPO        & 5e-6 & popular & 0.998 & 0.998 \\
NPO        & 2e-5 & rare    & 0.994 & 0.987 \\
NPO        & 4e-5 & rare    & 0.994 & 0.312 \\
NPO        & 5e-6 & rare    & 0.994 & 0.994 \\
\bottomrule
\end{tabular}%
}
\end{table}

\begin{table}[t]
\centering
\footnotesize
\caption{Full-parameter SFT with LoRA unlearning on the multi-domain split at $lr=2\times10^{-5}$. All algorithms show near-complete output collapse, in contrast to LoRA SFT (Table~\ref{tab:forget_retain_llama}).}
\label{tab:full_sft_multi}
\setlength{\tabcolsep}{4pt}
\resizebox{\columnwidth}{!}{%
\begin{tabular}{lllcc}
\toprule
Algorithm & LR & Popularity & ROUGE-L Before & ROUGE-L After \\
\midrule
GradAscent & 2e-5 & popular & 0.988 & 0.000 \\
GradAscent & 2e-5 & rare    & 0.964 & 0.000 \\
GradDiff   & 2e-5 & popular & 0.988 & 0.001 \\
GradDiff   & 2e-5 & rare    & 0.964 & 0.001 \\
NPO        & 2e-5 & popular & 0.988 & 0.144 \\
NPO        & 2e-5 & rare    & 0.964 & 0.046 \\
\bottomrule
\end{tabular}%
}
\end{table}

\subsection{Forget size effect.}
\label{sec:appendix_forget_size}
\begin{figure*}[t]
    \centering
    \includegraphics[width=\textwidth]{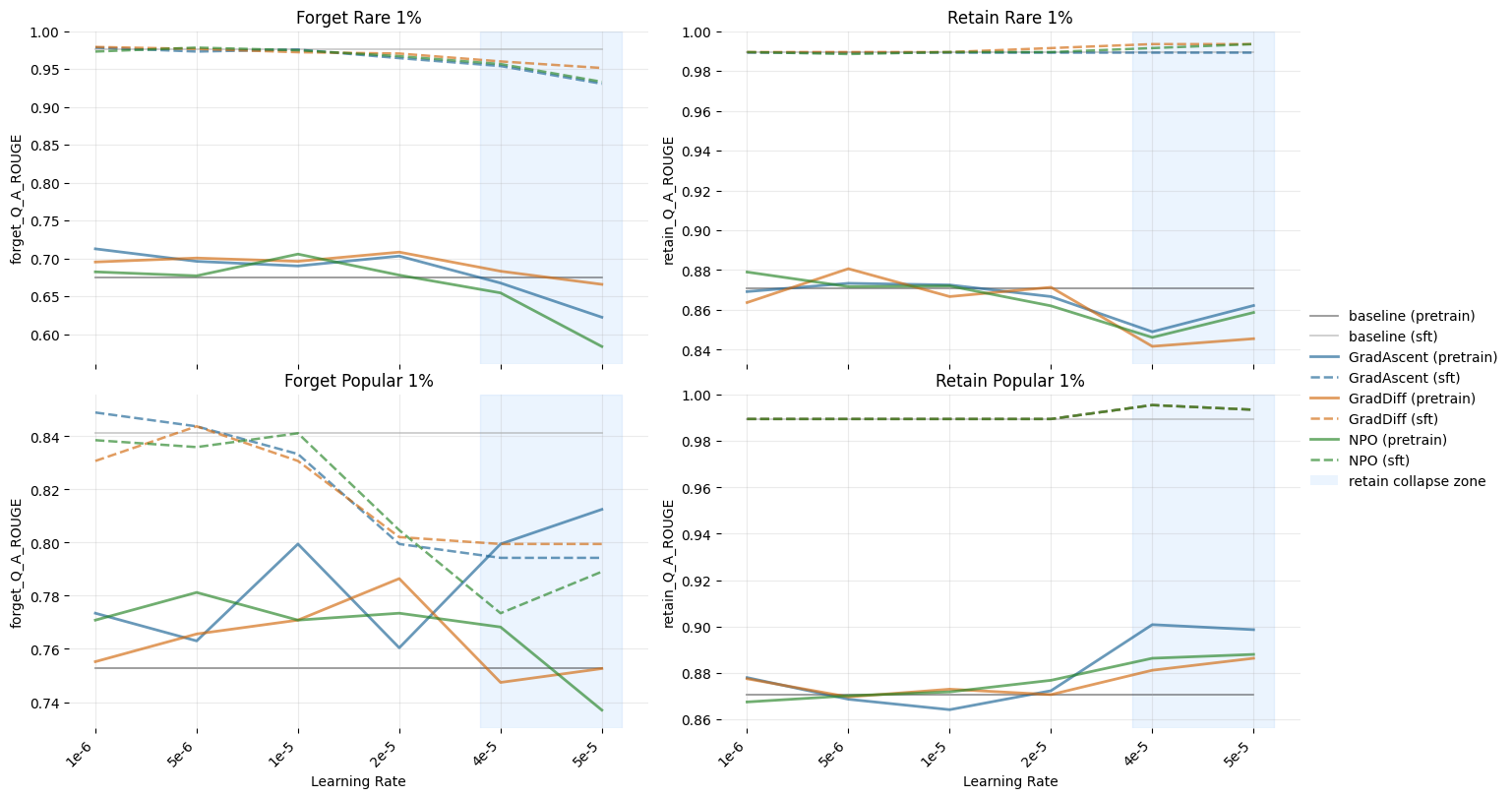}
    \caption{City, $N=1\%$. Top: rare; bottom: popular. Left: forget (\texttt{city\_forget\_\{rare,popular\}\_1}); right: retain (\texttt{city\_fast\_retain\_500}). ROUGE is reported across learning rates. Pretrain is shown as a solid line with circles; SFT as a dashed line with stars. Baselines are solid horizontal lines.}
    \label{fig:rare_forget_1}
\end{figure*}

\begin{figure*}[t]
    \centering
    \includegraphics[width=\textwidth]{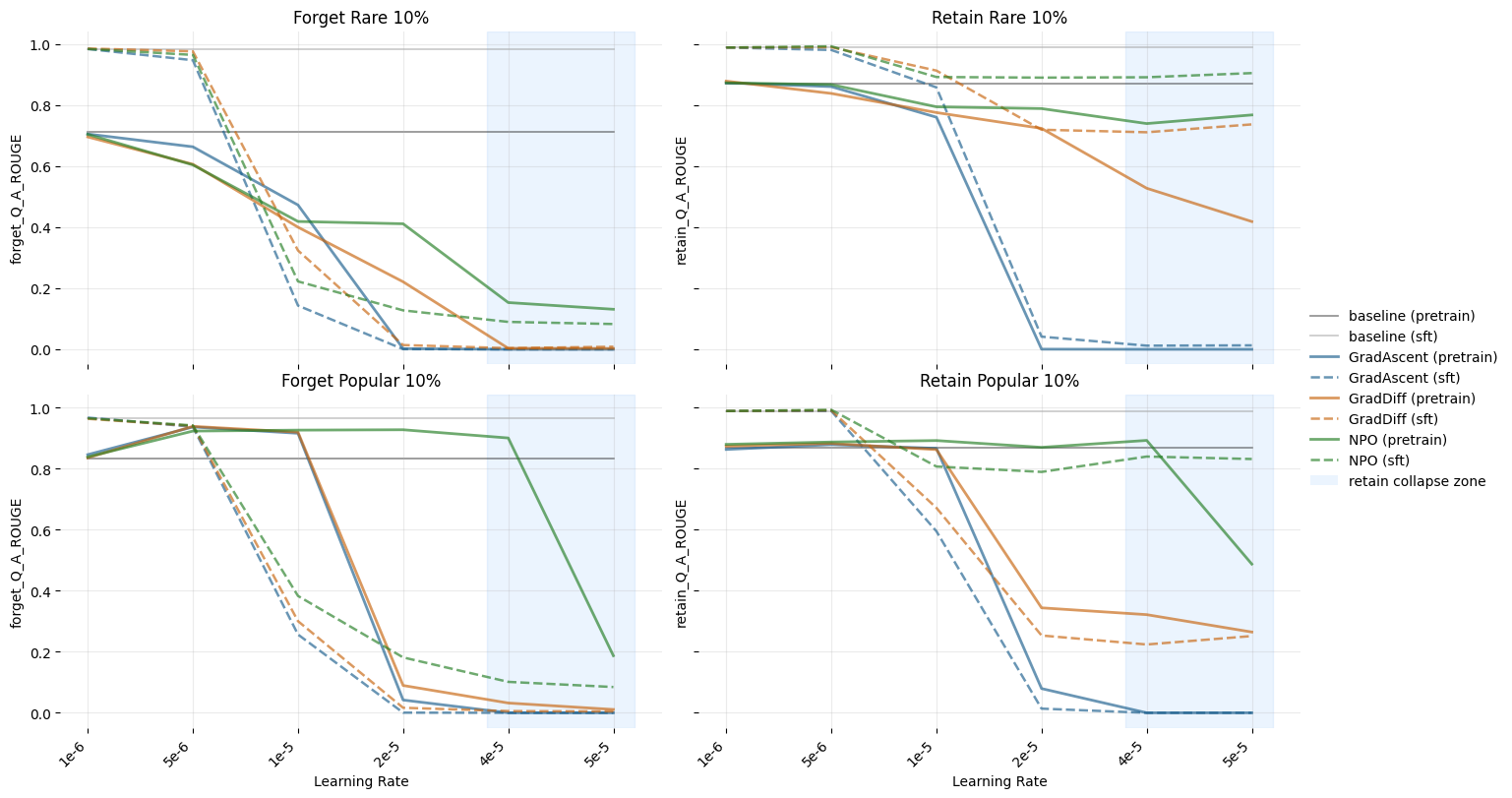}
    \caption{City, $N=10\%$. Top: rare; bottom: popular. Left: forget (\texttt{city\_forget\_\{rare,popular\}\_1}); right: retain (\texttt{city\_fast\_retain\_500}). ROUGE is reported across learning rates. Pretrain is shown as a solid line with circles; SFT as a dashed line with stars. Baselines are solid horizontal lines.}
    \label{fig:rare_forget_10}
\end{figure*}

Both \(N=1\%\) (Figure~\ref{fig:rare_forget_1}) and \(N=10\%\) (Figure~\ref{fig:rare_forget_10}) follow the same qualitative trends as \(N=5\%\) (Figure~\ref{fig:rare_forget_5}). At \(N=1\%\) the unlearning signal is weaker, so the forget set is harder to erase: ROUGE-L decreases more slowly and typically requires higher learning rates or additional epochs to match the \(N=5\%\) effect, while retention is comparatively stable. At \(N=10\%\) the signal is stronger, so forgetting progresses faster than at \(N=5\%\) and at lower learning rates, with collateral degradation on the retain set emerging earlier, especially for the Pretrained model. These tendencies hold for both rare and popular subsets.

\subsection{LLM as a Judge.}
\label{subsec:llm_judge}

Table~\ref{tab:forget_retain_gemma} reports Gemma-7B results combining ROUGE-L with an LLM-as-a-Judge evaluation (DeepSeek v1) that scores each answer on \textit{Accuracy} (factual alignment with the reference) and \textit{Fluency} (linguistic naturalness). LLaMA-3.1-8B results are in Table~\ref{tab:forget_retain_llama} in the main paper.

Judge Accuracy tracks ROUGE-L closely: configurations with lower forget ROUGE are consistently rated as less accurate, and the rare/popular asymmetry in automatic metrics is reflected in model-based judgments. Fluency remains stable across algorithms, confirming that observed drops are driven by factual removal rather than language degradation.

\begin{table}[t]
\centering
\footnotesize
\setlength{\tabcolsep}{3pt}
\renewcommand{\arraystretch}{1.12}
\resizebox{\columnwidth}{!}{%
\begin{tabular}{lll rr cc cc}
\toprule
& & & \multicolumn{2}{c}{\textbf{ROUGE-L}} & \multicolumn{2}{c}{\textbf{Judge Acc.}} & \multicolumn{2}{c}{\textbf{Judge Flu.}} \\
\cmidrule(lr){4-5}\cmidrule(lr){6-7}\cmidrule(lr){8-9}
\shortstack{\textbf{Train}\\\textbf{type}} & \textbf{Algo} & \textbf{Pop.} &
Forget $\downarrow$ & Retain $\uparrow$ & Forget $\downarrow$ & Retain $\uparrow$ & Forget $\uparrow$ & Retain $\uparrow$ \\
\midrule
Pretrain & w/o & rare & 0.515 & 0.445 & 0.212 & 0.562 & 0.973 & 0.990 \\
\rowcolor{gray!10} SFT & w/o & rare & 0.973 & 0.923 & 0.770 & 0.903 & 0.952 & 0.915 \\
Pretrain & GA  & rare & 0.267 {\scriptsize \textcolor{green!50!black}{($-$0.25)}} & 0.148 {\scriptsize \textcolor{red!65!black}{($-$0.30)}} & 0.369 {\scriptsize \textcolor{red!65!black}{($+$0.16)}} & 0.269 {\scriptsize \textcolor{red!65!black}{($-$0.29)}} & 0.546 {\scriptsize \textcolor{red!65!black}{($-$0.43)}} & 0.830 {\scriptsize \textcolor{red!65!black}{($-$0.16)}} \\
\rowcolor{gray!10} SFT & GA  & rare & 0.337 {\scriptsize \textcolor{green!50!black}{($-$0.64)}} & 0.869 {\scriptsize \textcolor{red!65!black}{($-$0.05)}} & 0.316 {\scriptsize \textcolor{green!55!black}{($-$0.45)}} & 0.801 {\scriptsize \textcolor{red!65!black}{($-$0.10)}} & 0.765 {\scriptsize \textcolor{red!65!black}{($-$0.19)}} & 0.820 {\scriptsize \textcolor{red!65!black}{($-$0.10)}} \\
Pretrain & GD  & rare & 0.421 {\scriptsize \textcolor{green!50!black}{($-$0.09)}} & 0.633 {\scriptsize \textcolor{green!50!black}{($+$0.19)}} & 0.149 {\scriptsize \textcolor{green!55!black}{($-$0.06)}} & 0.645 {\scriptsize \textcolor{green!55!black}{($+$0.08)}} & 0.620 {\scriptsize \textcolor{red!65!black}{($-$0.35)}} & 0.877 {\scriptsize \textcolor{red!65!black}{($-$0.11)}} \\
\rowcolor{gray!10} SFT & GD  & rare & 0.342 {\scriptsize \textcolor{green!50!black}{($-$0.63)}} & 0.893 {\scriptsize \textcolor{red!65!black}{($-$0.03)}} & 0.325 {\scriptsize \textcolor{green!55!black}{($-$0.44)}} & 0.843 {\scriptsize \textcolor{red!65!black}{($-$0.06)}} & 0.823 {\scriptsize \textcolor{red!65!black}{($-$0.13)}} & 0.839 {\scriptsize \textcolor{red!65!black}{($-$0.08)}} \\
Pretrain & NPO & rare & 0.472 {\scriptsize \textcolor{green!50!black}{($-$0.04)}} & 0.643 {\scriptsize \textcolor{green!50!black}{($+$0.20)}} & 0.291 {\scriptsize \textcolor{red!65!black}{($+$0.08)}} & 0.635 {\scriptsize \textcolor{green!55!black}{($+$0.07)}} & 0.749 {\scriptsize \textcolor{red!65!black}{($-$0.22)}} & 0.909 {\scriptsize \textcolor{red!65!black}{($-$0.08)}} \\
\rowcolor{gray!10} SFT & NPO & rare & 0.298 {\scriptsize \textcolor{green!50!black}{($-$0.68)}} & 0.873 {\scriptsize \textcolor{red!65!black}{($-$0.05)}} & 0.318 {\scriptsize \textcolor{green!55!black}{($-$0.45)}} & 0.818 {\scriptsize \textcolor{red!65!black}{($-$0.08)}} & 0.810 {\scriptsize \textcolor{red!65!black}{($-$0.14)}} & 0.822 {\scriptsize \textcolor{red!65!black}{($-$0.09)}} \\
\midrule
Pretrain & w/o & pop & 0.698 & 0.445 & 0.561 & 0.562 & 0.992 & 0.990 \\
\rowcolor{gray!10} SFT & w/o & pop & 0.994 & 0.923 & 0.907 & 0.903 & 0.932 & 0.915 \\
Pretrain & GA  & pop & 0.032 {\scriptsize \textcolor{green!50!black}{($-$0.67)}} & 0.110 {\scriptsize \textcolor{red!65!black}{($-$0.34)}} & 0.148 {\scriptsize \textcolor{green!55!black}{($-$0.41)}} & 0.224 {\scriptsize \textcolor{red!65!black}{($-$0.34)}} & 0.747 {\scriptsize \textcolor{red!65!black}{($-$0.24)}} & 0.838 {\scriptsize \textcolor{red!65!black}{($-$0.15)}} \\
\rowcolor{gray!10} SFT & GA  & pop & 0.000 {\scriptsize \textcolor{green!50!black}{($-$0.99)}} & 0.315 {\scriptsize \textcolor{red!65!black}{($-$0.61)}} & 0.190 {\scriptsize \textcolor{green!55!black}{($-$0.72)}} & 0.522 {\scriptsize \textcolor{red!65!black}{($-$0.38)}} & 0.297 {\scriptsize \textcolor{red!65!black}{($-$0.64)}} & 0.606 {\scriptsize \textcolor{red!65!black}{($-$0.31)}} \\
Pretrain & GD  & pop & 0.141 {\scriptsize \textcolor{green!50!black}{($-$0.56)}} & 0.242 {\scriptsize \textcolor{red!65!black}{($-$0.20)}} & 0.208 {\scriptsize \textcolor{green!55!black}{($-$0.35)}} & 0.408 {\scriptsize \textcolor{red!65!black}{($-$0.15)}} & 0.880 {\scriptsize \textcolor{red!65!black}{($-$0.11)}} & 0.853 {\scriptsize \textcolor{red!65!black}{($-$0.14)}} \\
\rowcolor{gray!10} SFT & GD  & pop & 0.965 {\scriptsize \textcolor{green!50!black}{($-$0.03)}} & 0.914 {\scriptsize \textcolor{red!65!black}{($-$0.01)}} & 0.886 {\scriptsize \textcolor{red!65!black}{($-$0.02)}} & 0.864 {\scriptsize \textcolor{red!65!black}{($-$0.04)}} & 0.806 {\scriptsize \textcolor{red!65!black}{($-$0.13)}} & 0.853 {\scriptsize \textcolor{red!65!black}{($-$0.06)}} \\
Pretrain & NPO & pop & 0.429 {\scriptsize \textcolor{green!50!black}{($-$0.27)}} & 0.533 {\scriptsize \textcolor{green!50!black}{($+$0.09)}} & 0.522 {\scriptsize \textcolor{green!55!black}{($-$0.04)}} & 0.574 {\scriptsize \textcolor{green!55!black}{($+$0.01)}} & 0.892 {\scriptsize \textcolor{red!65!black}{($-$0.10)}} & 0.893 {\scriptsize \textcolor{red!65!black}{($-$0.10)}} \\
\rowcolor{gray!10} SFT & NPO & pop & 0.961 {\scriptsize \textcolor{green!50!black}{($-$0.03)}} & 0.931 {\scriptsize \textcolor{green!50!black}{($+$0.01)}} & 0.887 {\scriptsize \textcolor{red!65!black}{($-$0.02)}} & 0.877 {\scriptsize \textcolor{red!65!black}{($-$0.03)}} & 0.893 {\scriptsize \textcolor{red!65!black}{($-$0.04)}} & 0.827 {\scriptsize \textcolor{red!65!black}{($-$0.09)}} \\
\bottomrule
\end{tabular}%
}
\caption{Gemma-7B at $N=5\%$, $lr=2\times10^{-5}$. ROUGE-L measures lexical forgetting and retention. Judge Accuracy measures factual alignment; Judge Fluency measures linguistic naturalness (DeepSeek v1). Parentheses show change from the matching \textit{w/o} baseline.}
\label{tab:forget_retain_gemma}
\end{table}

\subsection{General LLM Benchmarks Evaluation.}
\label{subsec:llm_benchs}

\begin{figure*}[t]
    \centering
    \includegraphics[width=\textwidth]{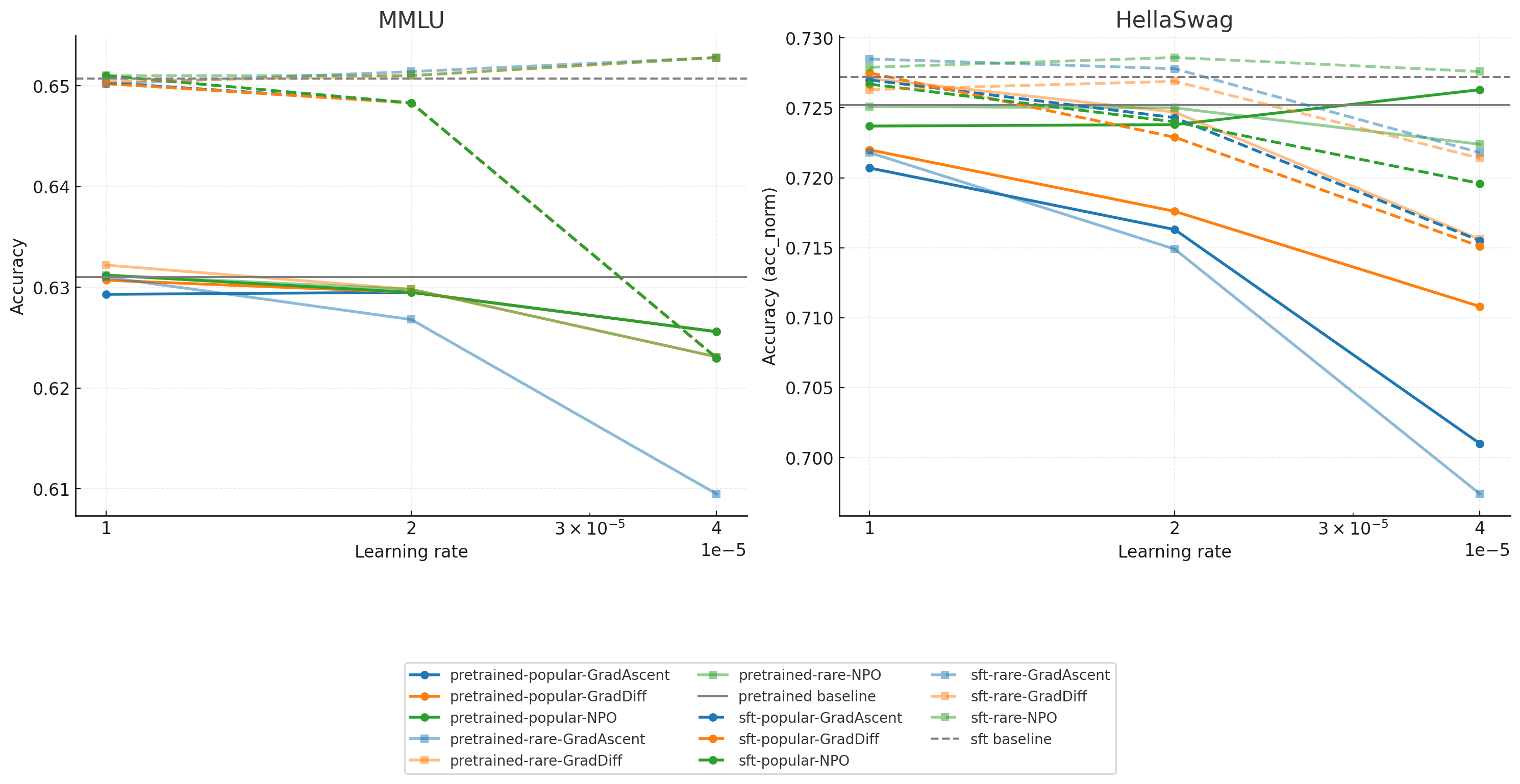}
    \caption{Evaluation on LLM benchmarks. Left: MMLU accuracy. Right: HellaSwag accuracy (acc\_norm). Curves show unlearning algorithms (GradAscent, GradDiff, NPO) on rare and popular forget sets across learning rates; pretrained models are solid, SFT models dashed; horizontal lines indicate the corresponding baselines.}
    \label{fig:llm_benchs}
\end{figure*}

Figure~\ref{fig:llm_benchs} shows that all three unlearning algorithms (GradAscent, GradDiff, NPO) preserve general model capabilities throughout the unlearning process. Across both rare and popular forget sets and all tested learning rates, deviations from the pretrained or SFT baseline on MMLU and HellaSwag remain within 3\%. Neither metric exhibits a consistent downward trend as the learning rate increases, confirming that the forgetting signal does not propagate to general knowledge. Pretrained and SFT variants behave similarly on these benchmarks, which stands in contrast to their divergent behavior on the task-specific retain set, further supporting the conclusion that capability degradation during unlearning is localized to the domain being unlearned rather than being a general phenomenon.

\subsection{Intrinsic memorization analysis}
\label{subsec:intrinsic}

We analyze intrinsic memorization signals to understand how rare and popular facts are internally represented and how unlearning affects them. We focus on two complementary diagnostics: (i) token-level probability and rank shifts of the gold answer, and (ii) hidden-state similarity across model layers. All analyses contrast pretrained and SFT models and explicitly separate rare and popular facts.

\paragraph{Token probability and rank shifts.}
We measure changes in the conditional probability $\log P_\theta(a \mid q)$ of the gold answer token and its rank in the output distribution. Table~\ref{tab:logp_rank_combined} reports results for both GradAscent (GA) and NPO.

Across both algorithms, unlearning popular facts leads to larger absolute probability shifts than rare facts. However, rank dynamics reveal a qualitatively different behavior: unlearning rare facts induces substantially larger rank changes than popular ones. This effect is most pronounced in the pretrained model, where forgetting rare facts shifts the average rank from single digits to above 200, while popular facts exhibit only mild rank perturbations. In SFT models, probability shifts remain non-trivial, but rank changes become more controlled and less extreme.

Rank-based diagnostics reveal a strong asymmetry between rare and popular facts that is not captured by changes in probability alone. Rare facts are significantly more fragile under unlearning, particularly in pretrained models, while SFT stabilizes rank behavior.

\begin{table}[t]
\centering
\footnotesize
\caption{Token-level probability and rank shifts under unlearning for GradAscent (GA) and NPO. Values are averaged over samples; $\Delta$ denotes unlearned minus base.}
\label{tab:logp_rank_combined}
\resizebox{\columnwidth}{!}{%
\begin{tabular}{lllc ccc ccc}
\toprule
Alg. & Set & Phase & Pop. & $n$ &
$\Delta \log P$ &
rank$_{\text{base}}$ &
rank$_{\text{unl}}$ &
$\Delta$rank \\
\midrule
GA  & forget & pretrain & popular & 482 & 3.88 & 34.2 & 41.8 & 7.5 \\
GA  & forget & pretrain & rare    & 482 & 2.07 & 8.8  & 230.4 & 221.6 \\
GA  & forget & sft      & popular & 482 & 10.94 & 59.9 & 55.2 & -4.6 \\
GA  & forget & sft      & rare    & 482 & 2.72 & 111.3 & 49.7 & -61.6 \\
NPO & forget & pretrain & popular & 482 & 0.32 & 34.2 & 35.3 & 1.1 \\
NPO & forget & pretrain & rare    & 482 & 0.99 & 8.8  & 17.6 & 8.9 \\
NPO & forget & sft      & popular & 482 & 7.99 & 59.9 & 56.0 & -3.9 \\
NPO & forget & sft      & rare    & 482 & 2.26 & 111.3 & 43.3 & -68.0 \\
\bottomrule
\end{tabular}%
}
\end{table}

\paragraph{Hidden-state similarity.}
We further analyze how unlearning modifies internal representations by computing hidden-state similarity between base and unlearned models across all layers. Table~\ref{tab:hidden_similarity_combined} summarizes cosine similarity and $\ell_2$ distance statistics for both GA and NPO.

In pretrained models, hidden states remain highly similar even for facts targeted by unlearning, indicating limited internal adaptation. In contrast, SFT models exhibit well-localized changes in representation. When forgetting popular facts, internal states shift substantially, while rare facts remain largely intact; the reverse occurs when forgetting rare facts. This suggests that SFT enables more selective and controlled modification of internal representations.

\textit{Conclusion.}
SFT produces sharper and more localized forgetting: hidden states shift substantially for the targeted fact type while remaining stable for the other, whereas pretrained models show limited internal adaptation regardless of which facts are unlearned.

\begin{table}[t]
\centering
\footnotesize
\caption{Hidden-state similarity between base and unlearned models for GradAscent (GA) and NPO. Metrics are averaged across layers and samples.}
\label{tab:hidden_similarity_combined}
\resizebox{\columnwidth}{!}{%
\begin{tabular}{lllc cc cc}
\toprule
Alg. & Phase & Unlearn & Pop. &
cosine &
$\ell_2$ &
cosine (std) &
$\ell_2$ (std) \\
\midrule
GA  & pretrain & pop  & popular & 0.982 & 3.35 & 0.006 & 0.61 \\
GA  & pretrain & rare & rare    & 0.981 & 2.70 & 0.008 & 0.66 \\
GA  & sft      & pop  & popular & 0.958 & 5.47 & 0.016 & 1.01 \\
GA  & sft      & rare & rare    & 0.966 & 3.81 & 0.011 & 0.81 \\
NPO & pretrain & pop  & popular & 0.996 & 1.52 & 0.001 & 0.27 \\
NPO & pretrain & rare & rare    & 0.990 & 1.98 & 0.003 & 0.36 \\
NPO & sft      & pop  & popular & 0.967 & 4.82 & 0.014 & 0.98 \\
NPO & sft      & rare & rare    & 0.973 & 3.40 & 0.009 & 0.72 \\
\bottomrule
\end{tabular}%
}
\end{table}

Taken together, token-level and hidden-state diagnostics confirm that rare and popular facts respond asymmetrically to unlearning at the representation level. SFT substantially improves the selectivity of forgetting: it concentrates representation changes on the targeted fact type while leaving the other largely intact, an effect absent in pretrained models.


\subsection{Distillation-based unlearning (UNDIAL)}
\label{sec:appendix_undial}

To broaden validation beyond SOTA unlearning methods, we also ran a distillation-based forgetting algorithm, Self-Distillation with Adjusted Logits (UNDIAL)~\cite{dong2025undial}. Unlike gradient updates that directly modify model parameters to suppress target knowledge, UNDIAL performs controlled self-distillation at the token level, aiming to adjust the model's output distribution to reduce reliance on the targeted content while preserving general capabilities. We evaluated UNDIAL on DUET for both \texttt{popular} and \texttt{rare} forget splits, with pretrained and SFT checkpoints and across multiple learning rates (Table~\ref{tab:undial_appendix}).

Overall, UNDIAL exhibits the same qualitative pattern we observe throughout the paper. For SFT models, retention remains very high across learning rates (retain ROUGE $\approx 0.987$ to $1.000$), while forgetting behaves more smoothly and predictably, especially on the popular split where forget ROUGE decreases as the learning rate increases. In contrast, pretrained models show substantially weaker retention (retain ROUGE $\approx 0.879-0.920$) and less stable forgetting dynamics across learning rates, with popular forgetting not improving monotonically. Importantly, the rare vs.\ popular asymmetry persists under self-distillation, suggesting that the popularity effect is not tied to a specific unlearning mechanism, but reflects a more general interaction between fact popularity and the model's training stage.

\begin{table}[t]
\centering
\footnotesize
\caption{UNDIAL (self-distillation) unlearning results on DUET across learning rates, evaluated on popular and rare forget splits.}
\label{tab:undial_appendix}
\setlength{\tabcolsep}{4pt}
\resizebox{\columnwidth}{!}{%
\begin{tabular}{lllc cc}
\toprule
Train type & Algorithm & Forget set & LR &
Forget & Retain \\
\midrule
SFT  & UNDIAL & popular & 1e-6 & 0.931 & 0.990 \\
SFT  & UNDIAL & popular & 2e-5 & 0.865 & 0.998 \\
SFT  & UNDIAL & popular & 5e-5 & 0.823 & 1.000 \\
SFT  & UNDIAL & rare    & 1e-6 & 0.988 & 0.990 \\
SFT  & UNDIAL & rare    & 2e-5 & 0.648 & 0.987 \\
SFT  & UNDIAL & rare    & 5e-5 & 0.679 & 0.994 \\
\midrule
pretrain & UNDIAL & popular & 1e-6 & 0.771 & 0.885 \\
pretrain & UNDIAL & popular & 2e-5 & 0.753 & 0.890 \\
pretrain & UNDIAL & popular & 5e-5 & 0.780 & 0.920 \\
pretrain & UNDIAL & rare    & 1e-6 & 0.694 & 0.879 \\
pretrain & UNDIAL & rare    & 2e-5 & 0.562 & 0.884 \\
pretrain & UNDIAL & rare    & 5e-5 & 0.608 & 0.922 \\
\bottomrule
\end{tabular}%
}
\end{table}

\textit{Conclusion.}
UNDIAL results are consistent with the gradient-based findings: SFT models produce more stable and effective forgetting, retain near-baseline knowledge on the retain set, and exhibit the same rare/popular asymmetry. The popularity effect thus generalizes across unlearning mechanisms with distinct optimization objectives.

\subsection{Model effect.}
\label{add_experiment}
\begin{figure*}[t]
    \centering
    \includegraphics[width=\textwidth]{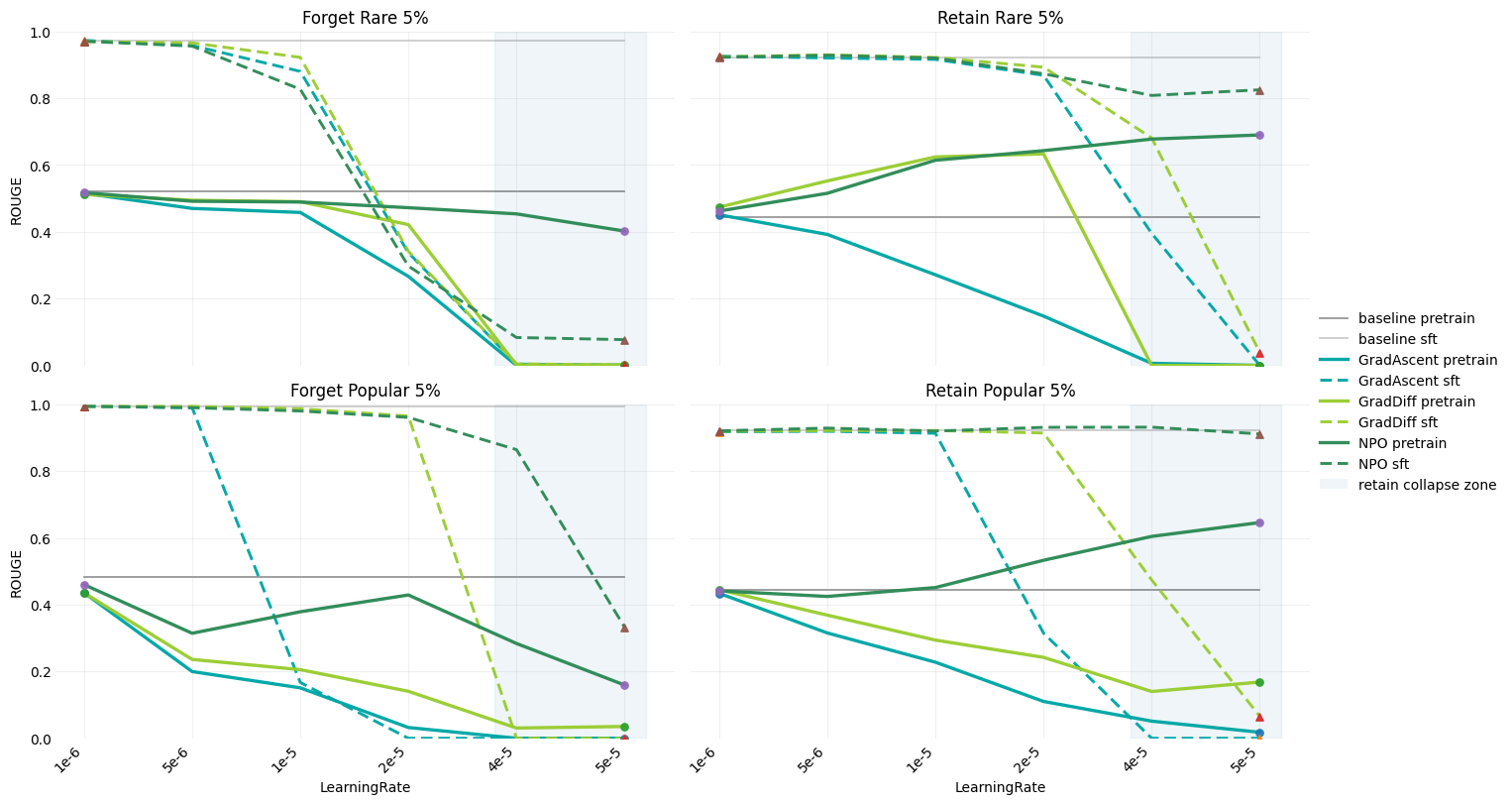}
    \caption{City split ($N=5\%$) evaluated on Gemma 7B. Top: rare; bottom: popular. Left: forget sets (\texttt{city\_forget\_\{rare,popular\}\_5}); right: retain set (\texttt{city\_fast\_retain\_500}). ROUGE scores are shown across learning rates. Pretrained models are plotted as solid lines with circles, and SFT models are plotted as dashed lines with stars. Solid horizontal lines indicate baselines.}
    \label{fig:llm_gemma}
\end{figure*}

For Gemma, we observe a consistent trend across forget sets (Figure \ref{fig:llm_gemma}): rare entities are easier to erase than popular ones, as shown by the smooth exponential decay of ROUGE-L with increasing learning rate. The SFT model, however, exhibits greater sensitivity to the learning rate, with performance dropping sharply from nearly 1.0 to 0 over a narrow range. NPO maintains stable retention across both model types, while GradAscent and GradDiff lead to considerable degradation, especially in the SFT variant (see Table~\ref{tab:forget_retain_gemma} in Appendix~\ref{subsec:llm_judge}). Overall, the SFT model preserves rare-entity knowledge more effectively than the pretrained model, but their response curves differ substantially. This divergence may reflect architectural differences between Gemma and LLaMA, as well as weaker alignment between Wikipedia-based popularity and Gemma's pretraining corpus.

\begin{figure*}[t]
    \centering
    \includegraphics[width=\textwidth]{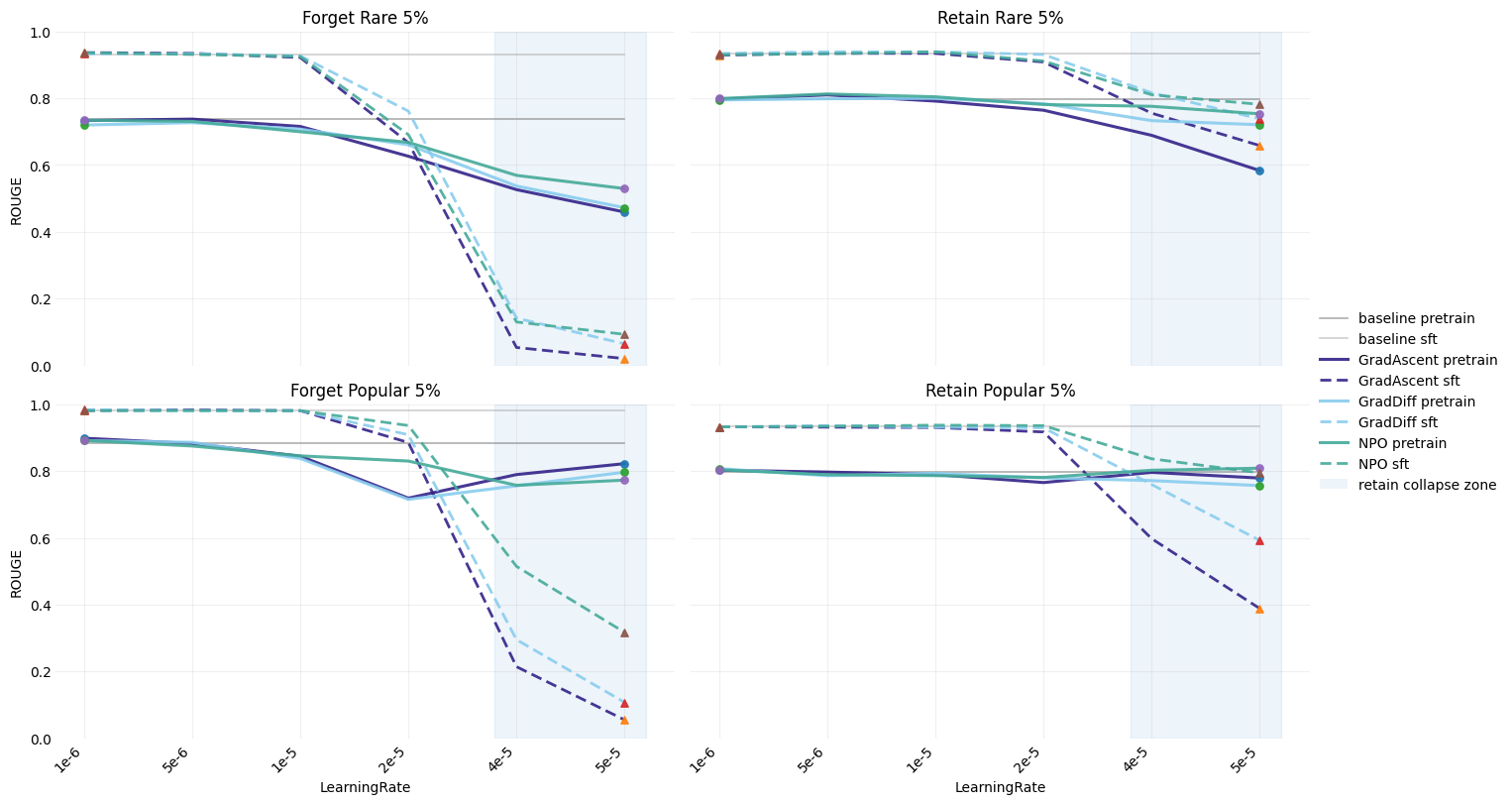}
    \caption{City split ($N=5\%$) evaluated on Qwen-2.5 7B. Top: rare; bottom: popular. Left: forget sets (\texttt{city\_forget\_\{rare,popular\}\_5}); right: retain set (\texttt{city\_fast\_retain\_500}). ROUGE scores are shown across learning rates. Pretrained models are plotted as solid lines with circles, and SFT models are plotted as dashed lines with stars. Solid horizontal lines indicate baselines.}
    \label{fig:llm_qwen}
\end{figure*}

For Qwen 2.5 7B in the Figure \ref{fig:llm_qwen}, the effect observed is more similar to the trends seen in LLaMA -3.1-8B.

Experiments on all three models show general trends, thereby generalizing and reinforcing the conclusions in the main part of the article.

\subsection{Multi-domain validation beyond cities.}
\label{subsec:multidomain}
While DUET is dominated by the \textit{city} domain, this choice was intentional to isolate popularity effects without noise in domain-induced confounders. To verify that our conclusions are not domain-specific, we additionally conducted multi-domain experiments on a domain-balanced subset of DUET, using 500 forget and 500 retain examples across domains. Table~\ref{tab:multidomain_validation} reports representative results at learning rate 2e-5.

Across all algorithms and training regimes, the same qualitative pattern persists. Forgetting popular facts consistently results in stronger degradation on the forget split compared to rare facts, while retention performance remains higher. This effect is especially clear in SFT models, where rare facts are better preserved during popular-forget training. In pretrained models, the gap between popular and rare forgetting is even larger, confirming that popularity-driven asymmetry generalizes beyond the city domain.

\begin{table}[ht!]
\centering
\footnotesize
\caption{Multi-domain unlearning results on DUET subset. Metrics are reported for learning rate $2\mathrm{e}{-5}$.}
\label{tab:multidomain_validation}
\resizebox{\columnwidth}{!}{%
\begin{tabular}{lllc ccc}
\toprule
Train type & Algorithm & Forget set & LR &
Forget & Retain \\
\midrule
SFT  & GradAscent & popular & 2e-5 & 0.461 & 0.736 \\
SFT  & GradAscent & rare    & 2e-5 & 0.651 & 0.729 \\
SFT  & GradDiff   & popular & 2e-5 & 0.477 & 0.743 \\
SFT  & GradDiff   & rare    & 2e-5 & 0.654 & 0.754 \\
SFT  & NPO        & popular & 2e-5 & 0.482 & 0.741 \\
SFT  & NPO        & rare    & 2e-5 & 0.668 & 0.731 \\
\midrule
pretrain  & GradAscent & popular & 2e-5 & 0.694 & 0.608 \\
pretrain  & GradAscent & rare    & 2e-5 & 0.527 & 0.485 \\
pretrain  & GradDiff   & popular & 2e-5 & 0.698 & 0.645 \\
pretrain  & GradDiff   & rare    & 2e-5 & 0.594 & 0.588 \\
pretrain  & NPO        & popular & 2e-5 & 0.715 & 0.686 \\
pretrain  & NPO        & rare    & 2e-5 & 0.602 & 0.578 \\
\bottomrule
\end{tabular}%
}
\end{table}

\textit{Conclusion.}
The rare/popular asymmetry is not an artifact of the city-heavy DUET distribution. It holds consistently across domains and training regimes, supporting the claim that fact popularity is a general determinant of unlearning difficulty.

\end{document}